\documentclass{article}

\PassOptionsToPackage{numbers, sort&compress}{natbib}
\usepackage[preprint]{neurips_2022}

\usepackage[utf8]{inputenc} 
\usepackage[T1]{fontenc}    
\usepackage{hyperref}       
\usepackage{url}            
\usepackage{booktabs}       
\usepackage{amsfonts}       
\usepackage{nicefrac}       
\usepackage{microtype}      
\usepackage[table]{xcolor}         

\usepackage{physics}
\usepackage{sty/siunitx}
\DeclareSIUnit\year{yr}
\AtBeginDocument{\renewcommand\qty\SI}

\definecolor{redA100}{HTML}{FF8A80}
\definecolor{redA200}{HTML}{FF5252}

\definecolor{blueA100}{HTML}{82B1FF}
\definecolor{blueA200}{HTML}{448AFF}

\definecolor{purpleA100}{HTML}{EA80FC}
\definecolor{purpleA200}{HTML}{E040FB}
\newcommand*{\Rb}{\mathbb{R}}

\newcommand*{\T}{^{\intercal}}
\newcommand*{\Ham}{\mathcal{H}}

\makeatletter
\DeclareFontEncoding{LS1}{}{}
\DeclareFontEncoding{LS2}{}{\noaccents@}
\DeclareFontSubstitution{LS1}{stix}{m}{n}
\DeclareFontSubstitution{LS2}{stix}{m}{n}
\makeatother

\DeclareMathAlphabet{\mathscrstix}{LS1}{stixscr}{m}{n}
\SetMathAlphabet{\mathscrstix}{bold}{LS1}{stixscr}{b}{n}
\DeclareMathAlphabet{\mathcalstix}{LS2}{stixcal}{m}{n}
\SetMathAlphabet{\mathcalstix}{bold}{LS2}{stixcal}{b}{n}

\DeclareFontFamily{U}{BOONDOX-calo}{\skewchar\font=45 }
\DeclareFontShape{U}{BOONDOX-calo}{m}{n}{
  <-> s*[1.05] BOONDOX-r-calo}{}
\DeclareFontShape{U}{BOONDOX-calo}{b}{n}{
  <-> s*[1.05] BOONDOX-b-calo}{}
\DeclareMathAlphabet{\mathcalboondox}{U}{BOONDOX-calo}{m}{n}
\SetMathAlphabet{\mathcalboondox}{bold}{U}{BOONDOX-calo}{b}{n}
\DeclareMathAlphabet{\mathbcalboondox}{U}{BOONDOX-calo}{b}{n}

\newcommand*\genq{\mathcalboondox{q}}
\newcommand*\genp{\mathcalboondox{p}}

\newcommand*\point{\textrm{point}}
\newcommand*\resid{\textrm{resid}}

\newcommand*\vq{\vb{q}}
\newcommand*\vp{\vb{p}}
\newcommand*\vR{\vb{R}}
\newcommand*\vPi{\vb{\Pi}}

\newcommand*\vJ{\vb{J}}

\newcommand*\Dt{\Delta t}

\newcommand*\SEThreeN{\mathsf{SE(3)}^{\otimes N}}

\newcommand*\KE{\textrm{KE}}
\newcommand*\PE{\textrm{PE}}
\newcommand*\asym{\textrm{asym}}

\newcommand*\force{\textrm{force}}
\newcommand*\LieT{\textrm{Lie $\mathrm{T}_2$}}
\newcommand*\BLieT{{\textbf{Lie $\bm{\mathrm{T}_2}$}}}

\newcommand*\SymSymbol{\textcolor{MaterialRedA200}{\bm{\mathcal{S}}}}
\newcommand*\LieSymbol{\textcolor{MaterialBlueA200}{\bm{\mathcal{L}}}}

\makeatletter
\DeclareRobustCommand\onedot{\futurelet\@let@token\@onedot}
\def\@onedot{\ifx\@let@token.\else.\null\fi\xspace}

\def\ie{\emph{i.e}\onedot}

\makeatother

\usepackage{mathtools}
\usepackage{tabularx}
\usepackage{framed}
\usepackage{bbm}
\usepackage{xspace}
\usepackage{color}

\usepackage{amssymb}
\usepackage{bm}
\usepackage{amsthm}
\usepackage{multirow}
\usepackage{wrapfig}
\usepackage{makecell}
\usepackage{xcolor-material}
\usepackage{marginfix}

\usepackage[capitalise]{cleveref}

\usepackage{algorithm}
\usepackage{algorithmic}

\usepackage{booktabs}
\usepackage{caption}

\usepackage{placeins}

\usepackage{cases}
\usepackage{empheq}

\usepackage[all=normal,paragraphs=tight,floats=tight,mathspacing=tight,wordspacing=tight,wordspacingfraction=0.96,charwidths=tight,leading=tight]{savetrees}

\makeatletter
\renewcommand{\normalsize}{%
  \@setfontsize\normalsize\@xpt\@xipt
  \abovedisplayskip      3\p@ \@plus 2\p@ \@minus 5\p@
  \abovedisplayshortskip \z@ \@plus 3\p@
  \belowdisplayskip      \abovedisplayskip
  \belowdisplayshortskip 4\p@ \@plus 3\p@ \@minus 3\p@
}
\normalsize

\makeatother

\usepackage{titlesec}

\crefname{appendix}{App}{Apps}

\hypersetup{
    colorlinks,
    linkcolor={blue!90!black},
    citecolor={blue!90!black},
    urlcolor={blue!95!black}
}

\title{
Data-driven discovery of non-Newtonian astronomy via learning non-Euclidean Hamiltonian
}
\author{Oswin So\thanks{Work was done while Oswin was at Georgia Tech.}\\%
Massachusetts Institute of Technology, USA\\%
\texttt{oswinso@mit.edu}\\%
\And
Gongjie Li\\%
Georgia Institute of Technology, USA\\%
\texttt{gongjie.li@physics.gatech.edu}\\%
\And
Evangelos A. Theodorou\\%
Georgia Institute of Technology, USA\\%
\texttt{evangelos.theodorou@gatech.edu}\\%
\And
Molei Tao\\%
Georgia Institute of Technology, USA\\%
\texttt{mtao@gatech.edu}
} 
\begin{document}

\maketitle


\vspace{-1.3\baselineskip}

\begin{abstract}
    Incorporating the Hamiltonian structure of physical dynamics into deep learning models provides a powerful way to improve the interpretability and prediction accuracy. While previous works are  mostly limited to the Euclidean spaces, their extension to the Lie group manifold is needed when rotations form a key component of the dynamics, such as the higher-order physics beyond simple point-mass dynamics for $N$-body celestial interactions. Moreover, the multiscale nature of these processes presents a challenge to existing methods as a long time horizon is required. By leveraging a symplectic Lie-group manifold preserving integrator, we present a method for data-driven discovery of non-Newtonian astronomy. Preliminary results show the importance of both these properties in training stability and prediction accuracy.
\end{abstract}


\vspace{-\baselineskip}

\section{Introduction} \label{sec:intro}


\begin{wrapfigure}[16]{r}{0.33\textwidth}
  \vspace{-9.0mm}%
  \centering
  \includegraphics[width=\linewidth]{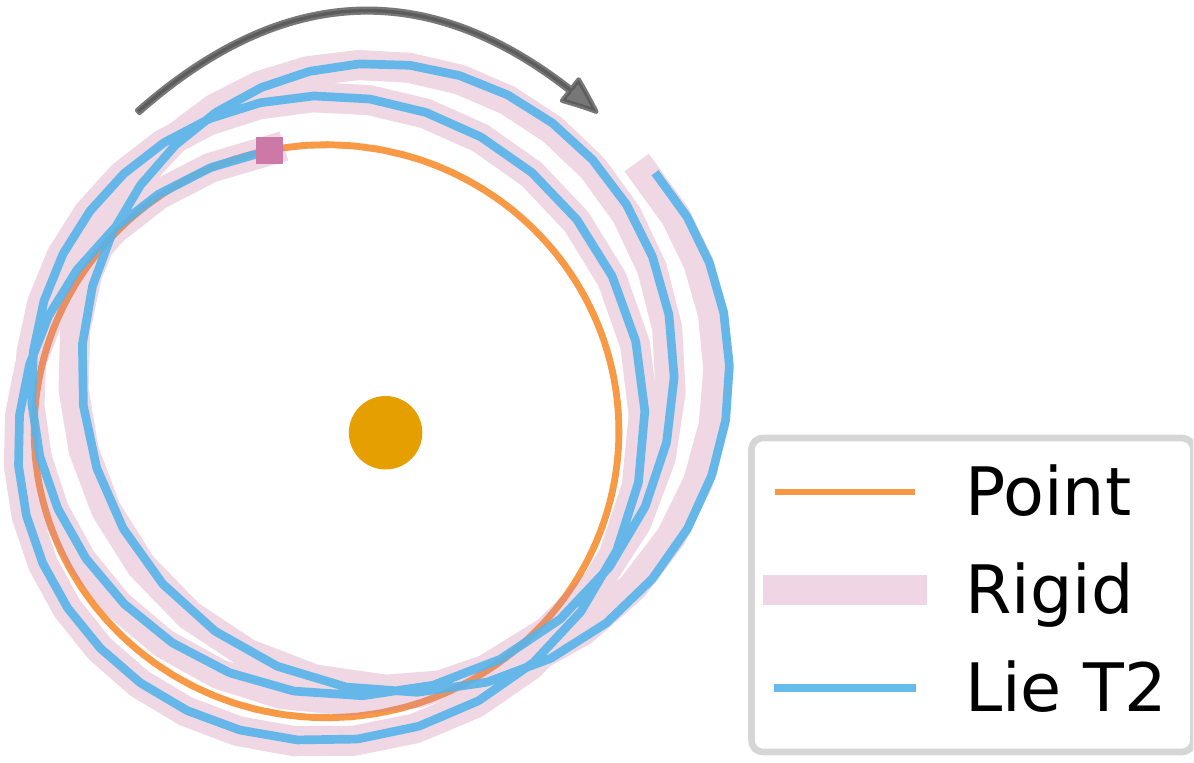}
  \captionof{figure}{One planet's orbit around a star: rigid body correction results in a \textit{precession}, \ie slow rotation of the orbital axis.
  Our method, `Lie T2', learns $V$ from data and predicts a trajectory that matches the ground truth with the rigid body potential.
  }
  \label{fig:toy:rigid_effect}
\end{wrapfigure}

Deep Neural Networks (DNN) have been demonstrated to be effective tools for learning dynamical systems from data.
One important class of systems to be learned have dynamics described by physical laws, whose structure can be exploited by learning the Hamiltonian of the system instead of the vector field \cite{greydanus2019hamiltonian, bertalan2019learning}.
An appropriately learned Hamiltonian can endow the learned system with properties such as superior long prediction accuracy \cite{chenDatadrivenprediction2021} and applicability to chaotic systems \cite{choudhary2020physics,chenDatadrivenprediction2021,han2021adaptable}.




To learn continuous dynamics from discrete data, one important step is to bridge the continuous and discrete times. Seminal work initially approximated the time derivative via finite differences and then matched it with a learned (Hamiltonian) vector field\cite{greydanus2019hamiltonian,bertalan2019learning}. Recent efforts avoid the inaccuracy of finite difference by numerically integrating the learned vector field. Especially relevant here is SRNN \cite{chen2019symplectic}, which uses a symplectic integrator to ensure the learned dynamics is symplectic (a necessity for Hamiltonian systems). Although 
SRNN only demonstrated learning separable Hamiltonians, breakthrough in symplectic integration of arbitrary Hamiltonians \cite{tao2016explicit} was used to extend SRNN \cite{xiong2020nonseparable}.
Further efforts on improving the time integration error have also been made \cite{dipietro2020sparse,david2021symplectic,mathiesen2022hyperverlet}.
Meanwhile, alternative approaches based on learning a symplectic map instead of the Hamiltonian also demonstrated efficacy \cite{jin2020sympnets,chenDatadrivenprediction2021}, although these approaches have not been extended to non-Euclidean problems.


In fact, one relatively under-explored area
is learning Hamiltonian dynamics on \textit{manifolds}
like the
Lie group manifold family\footnote{We note
extensions to
include holonomic constraints in \cite{finzi2020simplifying} and to handle contact in \cite{zhong2021extending}).}.
One important member of this family is $\mathsf{SO}(n)$, which describes isometries in $\Rb^n$ and is important for, e.g., dynamical astronomy.
The evolution of celestial bodies correspond to a mechanical system, and the 2- and 3-body problems have been a staple problem in works on learning Hamiltonian (e.g., \cite{greydanus2019hamiltonian,chen2019symplectic, chenDatadrivenprediction2021,lu2022modlanets}); however,
the Newtonian (point-mass) gravity considered is already well understood. Practical problems in planetary dynamics are complicated by higher-order physics such as planet spin-orbit interaction, tidal dissipation, and general relativistic correction.
While it is unclear what would be a perfect scientific model for these effects, 
planetary rotation is a necessary component to account for spin-orbit interaction and tidal forcing,
creating an $\mathsf{SO}(3)^{\otimes N}$ component of the configuration space.
To learn these physics from data, we need to learn on the Lie group.

Rigid body dynamics also play important roles in other applications such as robotics. In a seminal work \cite{duong2021hamiltonian}, Hamiltonian dynamics on $\mathsf{SO}(3)$ are used to learn rigid body dynamics for a quadrotor. In that work, Runge-Kutta 4 integrator is used. Consequently, the method is applicable to short time-horizon (see Sec.\ref{sec:results} and last paragraph of Sec.\ref{sec:2method}). 

For our problem of learning non-Newtonian astronomy, the time-horizon has to be long. Hence, we use a different approach by leveraging a Lie-group preserving symplectic integrator. 
Structure-preserving integration of dynamical systems on manifolds has been extensively studied in literature, for example for Lie groups \cite{iserles2000lie, bou2009hamilton,celledoni2014introduction,chen2021grit,celledoni2022lie} and more broadly, geometric integration \cite{haier2006geometric,leimkuhler2004simulating,blanes2017concise,sanz2018numerical}.


In summary, we propose a deep learning methodology for performing data-driven discovery of non-Newtonian astronomy. By leveraging the use of a symplectic Lie-group manifold preserving integrator, we show how a non-Euclidean Hamiltonian can be learned for accurate prediction of non-Newtonian effects. Moreover, we provide insights that show the importance of both symplecticity and exact preservation of the Lie-group manifold in training stability.


\section{Method}
\label{sec:2method}
Given observations of a dynamically evolving system, our goal is to learn the \textit{physics} that governs its evolution from the data.
Denote by $(\vq_{k, l}, \vR_{k, l}, \vp_{k, l}, \vPi_{k, l})_{k=1}^K, \; l \in [L]$ a dataset of snapshots of $L$ continuous-time trajectories of a system with $N$ interacting rigid bodies. That is,
\[
    (\vq_{k, l}, \vR_{k, l}, \vp_{k, l}, \vPi_{k, l}) = \big( \vq_l(k \Dt), \vR_l(k\Dt), \vp_l(k\Dt), \vPi_l(k\Dt) \big),
\]
where $\vq_l(t), \vR_l(t), \vp_l(t), \vPi_l(t)$ is a solution of some latent Hamiltonian ODE to be learned corresponding to mechanical dynamics on $T^* \SEThreeN$. $\Delta t$ is a (possibly large) observation timestep, $\vR\in \mathsf{SO}(3)^{\otimes N}$ is the rotational configuration of the $N$ rigid bodies, and $\vPi \in \mathfrak{so}(3)^{\otimes N}$ denotes each's angular momentum in their respective body frames.

Importantly, since the configuration space $\mathcal{Q}=\SEThreeN$ is not flat, the mechanical dynamics are not given by $\dot{\genq} = \pdv{\Ham}{\genp}, \dot{\genp} = -\pdv{\Ham}{\genq}$ for some Hamiltonian $\Ham$ that depends on the generalized coordinates $\genq \in \mathcal{Q}$ and generalized momentum $\genp \in T^*_\genq \mathcal{Q}$. Instead, the equations of motion can be derived via either Lagrange multipliers \cite{chen2021grit,hairerGeometricNumericalIntegration2006} or a Lie group variational principle \cite{chen2021grit,taeyoungleeliegroup2005}, which will be
\begin{subequations}
\noindent\begin{minipage}{.27\textwidth}
\begin{align}
    \dot{\vb{q}}_i &= \vb{p}_i / m_i, \label{eq:dyn:qdot} \\
    \dot{\vb{p}}_i &= -\pdv{V}{\vb{p}_i} + F_{\vp_i}  \label{eq:dyn:pdot}
\end{align}
\end{minipage}%
\noindent\begin{minipage}{.73\textwidth}
\begin{align}
    \dot{\vb{R}}_i &= \vb{R}_i \widehat{\vb{J}_i^{-1} \vb{\Pi}_i} \label{eq:dyn:Rdot} \\
    \dot{\vb{\Pi}}_i &= \vb{\Pi}_i \times \vb{J}_i^{-1} \vb{\Pi}_i 
        - \left( \vb{R}_i\T \pdv{V}{\vb{R}_i} - \left( \pdv{V}{\vb{R}_i} \right)\T \vb{R}_i \right)^\vee + F_{\vPi_i} \label{eq:dyn:Pidot}
\end{align}
\end{minipage}
\end{subequations}
assuming a physical Hamiltonian $\Ham(\vq,\vR,\vp,\vPi)=\sum_{i=1}^N \frac{1}{2} \vp_i^T \vp_i/m_i+\sum_i \frac{1}{2}\vPi^T_i \vJ_i^{-1} \vPi_i+V(\vq,\vR)$ that sums total (translation and rotational) kinetic energy and interaction potential $V$, where $m_i, \vJ_i$ denote the mass and inertial tensor of the $i$th body, and $F_{\vp}, F_{\vPi}$ are forcing terms to model nonconervative forces.
$\vPi_i \in \Rb^{3}$ is a vector, $\wedge$ is the map from $\Rb^3$ to $\textrm{Skew}_3$ and $\vee$ is its inverse (\cite{chen2021grit} for more details).
By learning the potential $V$, external forcing $F_{\vp}$ and torque $F_{\vPi}$, we can learn the physics of the system. 



\subsection{Machine Learning Challenges Posed by Dynamical Astronomy}
We study this setup because it helps answer scientific questions like: what physics governs the motions of celestial bodies, such as planets in a planetary system? The leading order physics is of course already well known, namely these bodies can be approximated by point masses that are interacting through a $1/r$ gravitational potential. However, planets are not point masses, and their rotations matter because they shape planetary climates \cite{quarles2022milankovitch, chen2022low} and even feedback to their orbits \cite{2019NatAs}. This already starts to alter $V$ even if one only considers classical gravity. For example, the gravitational potential $V$ for interacting bodies of finite sizes should be
$V(\vq, \vR) = \sum_{i < j} V_{i,j}$, where
\begin{equation} \label{eq:grav_potential}
    V_{i,j}(\vq, \vR)
    = \int_{\mathcal{B}_i} \int_{\mathcal{B}_j}
        -\frac{\mathcal{G} \rho(\vb{x}_i) \rho(\vb{x}_j)}{\norm{\vb{q}_i + \vb{R}_i \vb{x}_i - \vb{q}_j - \vb{R}_j \vb{x}_j}}
    \dd{\vb{x}_i} \dd{\vb{x}_j}
    \;\;\; = \;\;\; \underbrace{ -\frac{\mathcal{G} m_i m_j}{ \norm{\vb{q}_i - \vb{q}_j} } }_{V_{i, j, \point}}
    + \underbrace{ O\left( \frac{1}{\norm{q_i - q_j}^2} \right) }_{V_{i,j,\resid}}.
\end{equation}
Working with the full potential is complicated since 
$\mathcal{B}_i$ is not known and the integral is not analytically known.
Can we directly learn $V_{\resid}$ from time-series data?

Classical gravity (i.e. Newtonian physics) is not the only driver of planetary motion --- tidal forces and general relativity (GR) matter too. The former provides a dissipation mechanism and
plays critical roles in altering planetary orbits \cite{mardlingCalculatingTidal2002, naoz2011hot}; the latter doesn't need much explanation and has been demonstrated by, e.g., Mercury's precessing orbit \cite{clemence1947relativity}. Tidal forces depend on celestial bodies' rotations \cite{hut1981tidal} and thus is a function of both $\vq, \vR$.
GR's effects cannot be fully characterized with classical coordinates $\vq,\vR,\vp,\vPi$,
but post-Newtonian approximations based purely on these coordinates are popular \cite{blanchet2014gravitational}.
Can we learn both purely from data if we did not have theories for either?


In addition to the scientific questions, there are also significant machine learning challenges:

\noindent\textbf{Multiscale dynamics.} Rigid-body correction ($V_{\resid}$), tidal force, and
GR
correction are all much smaller forces compared to point-mass gravity. Consequently, their effects do not manifest until long time. Thus, one challenge for learning them is that the dynamical system exhibits different behaviors over \textbf{multiple timescales}. It is reasonable to require long time series data for the small effects to be learned; meanwhile, when observations are expensive to make, the observation time step $\Delta t$ can be much longer than the smallest timescales. Can we still learn the physics in this case? We will leverage symplectic integrator and its mild growth of error over long time \cite{hairerGeometricNumericalIntegration2006, tao2016explicit} to provide a positive answer.

\noindent \textbf{Respecting the Lie group manifold. }
However, even having a symplectic integrator is not enough because the position variable of the latent dynamics (i.e. truth) stays on $\SEThreeN$. If the integrated solution falls off this manifold such that $\vR\T \vR = I$ no longer holds, it is not only incorrect but likely misleading for the learning of $V(\vq,\vR)$. Popular integrators
such as forward Euler, Runge-Kutta 4 (RK4) and Leapfrog \cite{greydanus2019hamiltonian,chen2019symplectic,gruver2022deconstructing} unfortunately do no maintain the manifold structure.
%

\subsection{Learning with Lie Symplectic RNNs}
Our method can be viewed as a Lie-group generalization of the seminal work of SRNN \cite{chen2019symplectic},
where a good integrator that is both \textit{symplectic} and \textit{Lie-group preserving} is employed as a recurrent block. 

\noindent \textbf{Lie $\mathrm{T}_2$: A Symplectic Lie-Group Preserving Integrator. }
To construct an integrator that achieves both properties, we borrow from \cite{chen2021grit} the idea of Lie-group and symplecticity preserving splitting, and split our Hamiltonian as
$\Ham = \mathcal{H}_{\KE} + \mathcal{H}_{\PE} + \mathcal{H}_{\asym}$, which contains the axial-symmetric kinetic energy, potential energy and asymmetric kinetic energy correction terms.
This
enables computing the \textit{exact} integrators $\phi_t^{[\KE]}, \phi_t^{[\PE]}$ and $\phi_t^{[\asym]}$ (see \cref{app:integrator_details} for details).
We then construct a 2nd-order symplectic integrator \LieT{} by applying the Strang composition scheme. To account for non-conservative forces, the corresponding non-conservative momentum update $\phi^{[\force]}: (\vp, \vPi) \gets F(\vq, \vR, \vp, \vPi)$ is inserted in the middle of the composition \cite{chen2021grit}. This gives $\phi_h^{[\LieT]}$ for stepsize $h$ as
\begin{equation}
    \phi_h^{\LieT} \coloneqq \phi_{h/2}^{[\KE]} \, \circ \, \phi_{h/2}^{[\PE]} \, \circ \, \phi_{h/2}^{[\asym]} \, \circ \, \phi_{h}^{[\force]} \, \circ \, \phi_{h/2}^{[\asym]} \, \circ \phi_{h/2}^{[\PE]} \, \circ \, \phi_{h/2}^{[\KE]}
\end{equation}

\begin{figure}
    \vspace*{-6mm}
    \centering
    \includegraphics[width=0.65\textwidth]{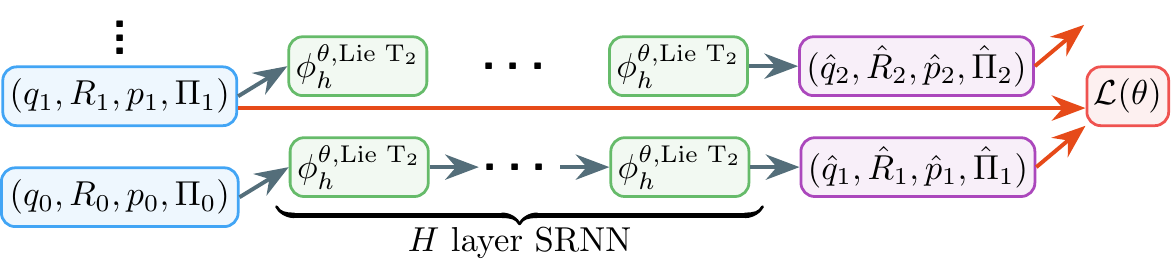}
    \caption{Inputs are fed through a recurrent layer with \LieT{}. Prediction error is used as a loss on $\theta$.}
    \label{fig:training_diagram}
    \vspace*{-4mm}
\end{figure}
\noindent \textbf{A Recurrent Architecture for Nonlinear Regression.}
Given the simplicity of $V_{\point}$, we assume this is known and learn $V_{\resid}$ and $F^\theta$
with multi-layer perceptron (MLP) $V^\theta_{\resid}$ and $F^\theta$ without assuming any pairwise structure (see \cref{app:training_details} for discussion). We then use
$\phi^{\theta, \LieT}$ to integrate dynamics forward, where $\theta$ denotes the dependence on the networks.
However, when the temporal spacing between observations $\Dt$ is large, using a single $\phi_{\Dt}^{\theta, \LieT{}}$ will result in large errors for the fast timescale dynamics.
Instead, we compose $\phi_h^{\theta, \LieT{}}$ $H$ times as $(\hat{\genq}_{k+1,l}, \hat{\genp}_{k+1,l}) = \phi_h^{\theta,\LieT} \circ \dots \circ \phi_h^{\theta,\LieT} ( \genq_{k,l}, \genp_{k,l} )$, where $H = h / \Dt \in \mathbb{Z}$ determines the integration stepsize $h$.
We perform training by minimizing the following empirical loss over random minibatches of size $N_b$
\begin{equation}
    \mathcal{L}(\theta) \coloneqq \frac{1}{N_b \, K} \sum_{l=1}^{N_b} \sum_{k=1}^{K} \Big\{
        \norm{\genq_{k, l} - \hat{\genq}_{k, l}^\theta}^2_2 + \norm{\genp_{k,l} - \hat{\genp}_{k, l}^\theta}^2_2 \Big\}
\end{equation}
Note that we do not assume access to the true derivatives $\dot{\genq}_{k, j}$ and $\dot{\genp}_{k, j}$ used in the loss function of some works \cite{greydanus2019hamiltonian,greydanus2022dissipative,cranmer2020lagrangian}. Our training process in summarized in \cref{fig:training_diagram} (see \cref{app:training_details} for details).

\noindent \textbf{Benefit.}
Learning an accurate $V^\theta, F^\theta$ requires accurate numerical simulation which also leads to a trainable model.
Without preservation of the manifold structure, training can lead to `shortcuts' outside the manifold that seemingly match the data but completely mislead the learning.
Symplecticity also plays a vital role in controlling the long time integration error ---
under reasonable conditions, a $p$th-order symplectic integrator has linear $\mathcal{O}(\Delta t h^p)$ error bound, whereas a $p$th-order nonsymplectic one
has an exponential
$\mathcal{O}(e^{C\Delta t} h^p)$ error bound \cite{hairerGeometricNumericalIntegration2006,tao2016explicit}.
While these bounds do not matter for small $\Delta t$, they are significant for multiscale problems where $\Delta t$ is macroscopic but $h$ is microscopic.
Consequently, improving error estimates for a nonsymplectic integrator by reducing $h$ makes the RNN exponentially deep --- this often renders training difficult \cite{pascanu2013difficulty} and is not desirable.


\section{Results} \label{sec:results}

\begin{wrapfigure}[18]{r}{0.45\textwidth}
    \vspace{-14.2mm}
    \centering
    \includegraphics[width=\linewidth]{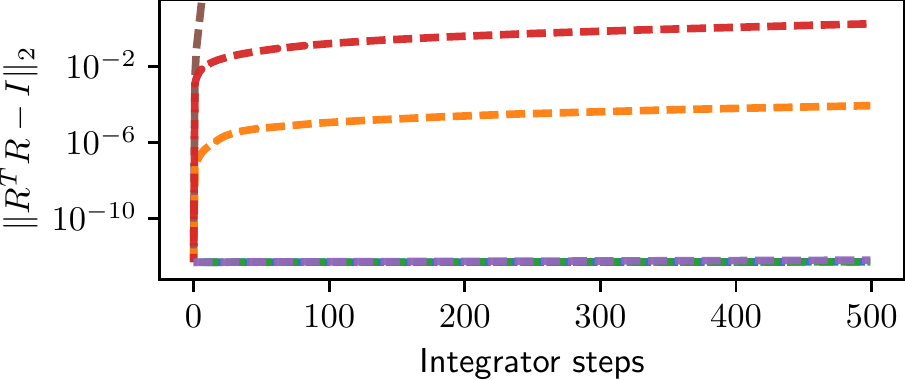}
    \hspace{0.005\textwidth}
    \includegraphics[width=\linewidth]{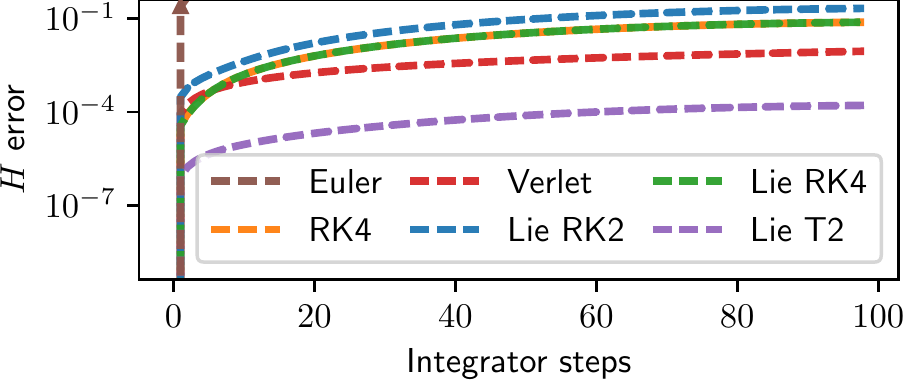}
    \caption{
        Results on TRAPPIST-1 with short data separation.
        \textbf{Top}: $SO(3)$ manifold error. \textbf{Bottom}: Hamiltonian error over the integrated trajectory.
        Only Lie T2 achieves low errors in both metrics.
        }
    \label{fig:trappist_easy:errs}
\end{wrapfigure}

We aim to answer two questions. \textbf{Q1} Can we learn multiscale physics? 
\textbf{Q2} How important are \textcolor{MaterialRedA200}{symplecticity} ($\SymSymbol$) and \textcolor{MaterialBlueA200}{Lie-group preservation} ($\LieSymbol$) for learning?
The closest baseline for our problem is
work of \cite{duong2021hamiltonian}, which learns short timescale rigid-body Hamiltonian dynamics for robotics. Placed in our framework, their work corresponds to using RK4 for the recurrent block,
which is neither $\SymSymbol$ nor $\LieSymbol$.
Therefore, to investigate \textbf{Q2}, we vary the choice of integrator in our framework as follows: 
\textit{Normal}: Explicit Euler, RK4.
$\SymSymbol$: Verlet. $\LieSymbol$: Lie RK2(CF2) and Lie RK4(CF4)
\cite{celledoni2022lie}. We leave the precise details to \cref{app:training_details,app:train_integrator_details}.

%
\textbf{Toy Two-Body Problem.}
We consider an illustrative two-body problem
to demonstrate the effects of $V_{\textrm{rigid}}$.
In \cref{fig:toy:rigid_effect} `Point' \& `Rigid' denote exact solutions for a point-mass and rigid-body potential,
and `Lie T2' the prediction of our method based on a $V$ learned from data.
Compared to `Point', `Rigid' induces an \textit{apsidal precession} (rotation of the orbital axis) due to spin-orbit couplings. Our method successfully predict this interaction and matches the trajectory of `Rigid'. 

We next test our method by learning the dynamics of the \texttt{TRAPPIST-1} system \cite{gillon2016temperate} which consists of seven earth-sized planets and is notable for potential habitability for terrestrial forms of lives.  


\textbf{TRAPPIST-1, Large $\Dt$.}
To answer \textbf{Q1}, we choose a large data timestep $\Delta t = \qty{2.4e-3}{\year}$. The closest planet has an orbital period of $\sim 2 \Delta t$ ($\qty{4.1e-3}{\year}$), while the rigid body correction, tidal force and GR correction act on much longer scales. Only \LieT{} successfully trains. All other methods diverge during training (denoted by $\infty$) despite attempts at stabilization with techniques such as normalization (LayerNorm \cite{ba2016layer}, GroupNorm \cite{wu2018group}).
Reducing $h$ improves integration accuracy, but increases the RNN depth and makes training more unstable.
We compare with the solution for point-mass potential only (No Correction). Our method reduces the error up to two orders of magnitude in measures of trajectory error and potential gradients (\cref{tab:trappist}). See \cref{app:eval_metrics} for column definitions.
\begin{table}
\vspace*{-1.2\baselineskip}
\newcommand*{\B}{\bfseries}%
\newcommand*{\NO}{$\phantom{\SymSymbol{}}\phantom{\LieSymbol{}}$}%
\newcommand*{\SymOnly}{$\SymSymbol{}\phantom{\LieSymbol{}}$}%
\newcommand*{\LieOnly}{$\phantom{\SymSymbol{}}\LieSymbol$}%
\newcommand*{\BothSym}{$\SymSymbol{}\LieSymbol$}%
\small
\centering
\setlength{\tabcolsep}{4pt}
\makebox[0pt][c]{\parbox{1.1\linewidth}{%
\begin{minipage}[b]{0.46\linewidth}\centering
    \caption{TRAPPIST-1 with long data separation evaluated after $500$ integrator steps. All methods except for $\LieT{}$ diverge during training.}
    \label{tab:trappist}
    \begin{tabular}{
        l
        <{\hspace{-1.6mm}}
        S[table-format=1.1e1,tight-spacing = true,detect-all]
        <{\hspace{-0.1mm}}
        S[table-format=1.1e1,tight-spacing = true,detect-all]
        <{\hspace{-0.1mm}}
        S[table-format=1.1e1,tight-spacing = true,detect-all]
        <{\hspace{-0.1mm}}
        S[table-format=1.1e1,tight-spacing = true,detect-all]
    }
        \toprule
        & $\norm{\Delta q}_{2}$
        & $\norm{\Delta R}$
        & $\norm{\Delta \dot{p}}_{2}$
        & $\norm*{\Delta \dot{\Pi}}_{2}$
        \\ \midrule
        No Corrections     & 8.6e-06 & 1.6e+00 & 3.3e+00 & 4.7e-02 \\ \addlinespace[\belowrulesep]
        \makecell{Euler, RK4, \\Verlet, Lie RK2, \\Lie RK4}  & $\infty$ & $\infty$ & $\infty$ & $\infty$ \\ \addlinespace[\belowrulesep]
        \rowcolor{purpleA100!30}\B \BLieT{} (Ours) & \B 6.1e-08 & \B 4.0e-02 & \B 1.2e-01 & \B 3.1e-03 \\
        \bottomrule
    \end{tabular}
\end{minipage}
\hfill
\begin{minipage}[b]{0.52\linewidth}\centering
    \caption{TRAPPIST-1 with short data separation evaluated after $500$ integrator steps.}
    \label{tab:trappist_easy}
    \begin{tabular}{
        l
        <{\hspace{-1.6mm}}
        S[table-format=1.1e1,tight-spacing = true,detect-all]
        <{\hspace{-0.1mm}}
        S[table-format=1.1e1,tight-spacing = true,detect-all]
        <{\hspace{-0.1mm}}
        S[table-format=1.1e1,tight-spacing = true,detect-all]
        <{\hspace{-0.7mm}}
        S[table-format=1.1e1,tight-spacing = true,detect-all]
    } 
        \toprule
        & $\norm{\Delta q}_{2}$
        & $\norm{\Delta R}$
        & $\norm*{\Delta \pdv{V}{\vq}}_{2}$
        & $\norm*{\Delta \pdv{V}{\vR}}_{2}$
        \\ \midrule
        \NO{}%
        Euler              & $\infty$ & $\infty$ & $\infty$ & $\infty$ \\
        \NO{}%
        RK4                & 1.6e-05 & 3.9e-01 & 1.5e+03 & 7.9e-01 \\
        \rowcolor{redA100!30}\SymOnly{}%
        Verlet             & 8.6e-08 & 2.0e+07 & \B 1.6e+01 & 4.1e-01 \\
        \rowcolor{blueA100!30}\LieOnly{}%
        Lie RK2            & 5.3e-05 & 3.3e-01 & 2.6e+02 & 7.6e-01 \\
        \rowcolor{blueA100!30}\LieOnly{}%
        Lie RK4            & 1.6e-05 & 3.5e-01 & 1.5e+03 & 8.4e-01 \\
        \rowcolor{purpleA100!30}\BothSym{}%
        \B Lie $\mathrm{T}_2$ (Ours) & \B 8.0e-08 & \B 2.4e-01 & \B 1.6e+01 & \B 4.0e-01 \\
        \bottomrule
    \end{tabular}
\end{minipage}
}}
\vspace*{-0.8\baselineskip}
\end{table}

\textbf{TRAPPIST-1, Small $\Dt$.}
To gain more insight on \textbf{Q2}, we shrink $\Dt$ until almost all methods can converge and only consider conservative forces (\ie no tidal force or GR). The mean errors in the predicted trajectory and derivatives of the learned potential $V$ after $500$ integrator steps are shown in \cref{tab:trappist_easy}. Both $\SymSymbol$ methods achieve small errors in position related terms. Verlet has a large rotational error since it does not integrate on the rotation manifold.
$\LieSymbol$ methods achieve lower rotational errors but are worse elsewhere. \LieT{} being both $\SymSymbol{}\LieSymbol{}$ achieves the lowest error on both fronts.

\newpage
\FloatBarrier
\section{Broader Impact} \label{sec:broader_impact}
Our work presents an approach for learning multiscale, higher order physics on the Lie-group manifold in the context of non-Newtonian astronomy. This research, though directly applicable to astronomy, can also be applied to perform data-driven discovery multiscale phenomena on Lie groups in other fields.

\bibliographystyle{unsrtnat} 
\bibliography{references.bib}

\newpage

\appendix
\input{subtex/appendix.tex}

\end{document}